\documentclass{article}
\usepackage{spconf,amsmath,graphicx}
\usepackage{amsfonts}
\usepackage[ruled,vlined]{algorithm2e}
\usepackage{booktabs}
\usepackage[flushleft]{threeparttable}
\SetKwComment{Comment}{$\triangleright$\ }{}

\newtheorem{theorem}{Theorem}[section]
\newtheorem{assumption}{Assumption}[section]

\title{Forgetting Private Textual Sequences in Language Models via Leave-One-Out Ensemble}

\name{Zhe Liu, Ozlem Kalinli}
\address{Meta AI, Menlo Park, CA, USA}

\begin{document}
\ninept
\maketitle
\begin{abstract}
Recent research has shown that language models have a tendency to memorize rare or unique token sequences in the training corpus. After deploying a model, practitioners might be asked to delete any personal information from the model by individuals’ requests. Re-training the underlying model every time individuals would like to practice their rights to be forgotten is computationally expensive. We employ a teacher-student framework and propose a novel leave-one-out ensemble method to unlearn the targeted textual sequences that need to be forgotten from the model. In our approach, multiple teachers are trained on disjoint sets; for each targeted sequence to be removed, we exclude the teacher trained on the set containing this sequence and aggregate the predictions from remaining teachers to provide supervision during fine-tuning. Experiments on LibriSpeech and WikiText-103 datasets show that the proposed method achieves superior privacy-utility trade-offs than other counterparts.
\end{abstract}
\begin{keywords}
Language models, memorization, ensemble methods, knowledge distillation, automatic speech recognition
\end{keywords}
\section{Introduction}
\label{sec:intro}

State-of-the-art language models (LMs) always involve training over large and diverse corpora which might include private information of individuals, such as names, addresses, and credit card numbers. Recent research has shown that such sensitive information in training datasets can be extracted by adversaries \cite{fredrikson2015model, song2019auditing, carlini2019secret, carlini2021extracting, huang2022large}. Specifically, LMs are prone to unintentionally \emph{memorize} rare and unique token sequences, and when being prompted appropriately, they could emit these memorized text verbatim \cite{carlini2022quantifying}. This is undesirable because such memorization may expose users' information.

Privacy risks persist even if adversaries do not have direct access to the deployed LMs. Automatic speech recognition (ASR) models are often accompanied by LMs via shallow fusion for boosting their overall quality \cite{kannan2018analysis, irie2019language}. It has been shown that target textual sequences in LM training data could still be detected when one has only black-box (query) access to LM-fused speech recognizer \cite{huang2022detecting}. Therefore, it has become crucial to provide privacy protections for LMs.

Modern data regulatory frameworks, such as the \emph{right to be forgotten} (RTBF), provide a mechanism under which users can request their data to be deleted \cite{mantelero2013eu}. Assuming any private textual sequences to be removed can be identifiable from the original training corpus, a naive approach is to re-train the underlying LMs \emph{every time} users want to practice their RTBF. However, this is not efficient and could be  computationally costly, especially for large LMs with billions of parameters. Thus, when practitioners are asked to delete private information from the LMs by users' requests, it calls for an efficient solution that can be applied
with just a few parameter updates to achieve the goal of forgetting targeted textual sequences with little to no degradation of general LM performance.

In this paper, we employ a teacher-student framework and propose a novel \emph{leave-one-out ensemble} method to \emph{unlearn} the targeted token sequences that are requested to be forgotten in the deployed LM. In our approach, in addition to the training of a base LM using the full dataset that contains user data, multiple teacher LMs are trained on disjoint training data from different users. Then when a request for deletion is received, the deployed base LM, the one who contains the sensitive information to be removed, is fine-tuned on targeted textual sequences with teachers' predictions being used as soft labels. Specifically, for each targeted sequence, we leave out the teacher that was trained on the set including this specific sequence, and then aggregate the predictions of remaining teachers to provide supervision during fine-tuning. Finally, the resulting student model is published to users as an update of LM.

Intuitively, given any targeted private sequence to be forgotten, by excluding the single ``vulnerable'' teacher, we expect its predicted probabilities from the aggregation of other benign teachers are close to these from the re-trained LM from scratch, with targeted sequence removed. Governed by the teacher ensembles, the fine-tuned LM is able to re-calibrate the probabilities on each of the private sequences upon requests and mitigate the memorization of them. By training for a few epochs, the overall LM accuracy shall be barely impacted.

We make the following contributions: (1) introducing the novel leave-one-out ensemble framework for effective unlearning of any targeted private information in LMs; (2) investigating the effect of several different knowledge distillation mechanisms for enhancing RTBF; (3) conducting theoretical analysis to measure the similarity of predicted probability distributions between leave-one-out teacher ensembles and the re-trained LM with all targeted private sequences removed; and (4) providing experimental results on comparing the privacy-utility trade-offs of various methods on LM and ASR tasks.


\section{Related Work}
\label{sec:related}
Prior work for mitigating privacy risks can be divided into privacy-preserving training methods and post-training unlearning methods. The former aims to bound the effect of any individual's data on the model's output, including federated learning (FL) \cite{konevcny2016federated2,mcmahan2017communication,ramaswamy2020training,thakkar2021understanding}, differential privacy (DP) \cite{dwork2006calibrating, dwork2014algorithmic}, and private aggregation of teacher ensembles (PATE) \cite{papernot2016semi,papernot2018scalable,liu2023mitigating}. In many cases, however, the cost of these methods is a reduction in the model's accuracy \cite{bagdasaryan2019differential, wang2019enhance}. More importantly, in practical scenarios, users may request to delete their personal information dynamically \emph{after} the model deployment, while these privacy-preserving training approaches are not intrinsically designed to handle the forgetting of \emph{targeted} token sequences. Thus, in the other line of work, post-training approaches have been investigated to avoid of the need for re-training the model \cite{bourtoule2021machine, liu2020federated, jagielski2022measuring, jang2022knowledge}. Particularly, authors in \cite{jang2022knowledge} propose gradient ascent method and show that it is effective at forgetting targeted token sequences.

\section{Methodology}
\label{sec:method}
\subsection{The Leave-One-Out Ensemble Framework}
In this section, we describe the leave-one-out ensemble framework. Specifically, subsections \ref{sec:method:pretrain} and \ref{sec:method:teacher} introduce the training of LM and a series of teacher models using private data of users; subsection \ref{sec:method:forget} illustrates the data removal requests; subsection \ref{sec:method:looe} develops a post-training unlearning approach which leads to a new model that can forget the targeted token sequences.
\subsubsection{Training Base LM on Private Data}
\label{sec:method:pretrain}
Let $\mathcal{D}_k=\mathcal{D}_k^{\text{pri}}\cup\mathcal{D}_k^{\text{pub}}$ be the LM training corpus from the $k$th user, where $\mathcal{D}_k^{\text{pri}}$ is a set of records with sensitive information and $\mathcal{D}_k^{\text{pub}}$ is a corpus without sensitive information. Then let $\mathcal{D}=\cup_{k=1}^K \mathcal{D}_k$ be the entire training set over all users' data.

Cross entropy (CE) loss is usually used for LM training. Given any training record $(w_1,w_2,\ldots,w_T) \in \mathcal{D}$, the following shows this function at step $t$
\begin{align}
\mathcal{L}^{\text{CE}}_t(\theta)=-\sum_{w\in V} \textbf{1}\{w=w_t\}\cdot\log p_{\theta}(w|w_{1:t-1})
\end{align}
where $\theta$ represents the model parameter, $V$ is the vocabulary set, and the LM predicts the word $w$ with a probability $p_{\theta}(w|w_{1:t-1})$ at step $t$. Suppose $\theta^{\text{base}}$ is an LM trained on $\mathcal{D}$ and we release it to users.

\subsubsection{Training Teacher Models on Private Data}
\label{sec:method:teacher}
In addition to the LM $\theta^{\text{base}}$ already trained on private data of $\mathcal{D}$, we train a series of teacher models, which will be utilized at some later time when users request that their information be deleted.

Each teacher is an LM trained independently on a subset of the entire training data. The data is partitioned into disjoint subsets by users to ensure no pair of teachers will have trained on data from the same user. That is, all records from any user is only included in the training corpus of one specific teacher model. This is the \emph{isolation} principal that any user's data is only exposed to one single teacher. 

Without the loss of generality, assuming the entire training set $\mathcal{D}$ is partitioned into $M$ disjoint subsets by users, denote $\mathcal{B}_{m}=\cup_{k=(m-1)d+1}^{md} \mathcal{D}_k$ for each $m=1,\ldots, M$ and $d=[K/M]$. Thus we have $\mathcal{D}=\cup_{m=1}^M \mathcal{B}_m$ and $\mathcal{B}_{m'}\cap\mathcal{B}_{m''}=\emptyset$ for any $m'\neq m''$. The $m$th teacher model is trained using data of $\mathcal{B}_m$.

Thus we obtain a set of trained teacher models $\{\theta_m\}_{m=1}^M$ along with the deployed LM $\theta^{\text{base}}$. These teachers are not released to users.

\subsubsection{Textual Sequences to Be Forgotten}
\label{sec:method:forget}
After the deployment of the original LM $\theta^{\text{base}}$, users might submit the requests to have some of their sensitive textual sequences deleted.

Denote $\mathcal{D}_k^{\text{forget}}$ as the set of token sequences to be forgotten requested by the $k$th user. Here, $\mathcal{D}_k^{\text{forget}}$ can be a subset of $ \mathcal{D}_k^{\text{pri}}$, while it is also possible that only a subsequence of tokens in any textual sentence is requested to be removed. For example, suppose the text of ``\ldots \text{ enter the code 1234 to unlock the door }\ldots'' is one training record contained in $\mathcal{D}_k^{\text{pri}}$, and only the numbers are to be forgotten, that is, ``1234'' $\in \mathcal{D}_k^{\text{forget}}$. Let $\mathcal{D}^{\text{forget}}=\cup_{k=1}^K \mathcal{D}_k^{\text{forget}}$.

One naive approach is to follow subsection \ref{sec:method:pretrain} and re-train the LM on the training corpus of $\mathcal{D}\,\backslash\,\mathcal{D}^{\text{forget}}$. We denote this resulting model as $\theta^{*}$. However, this could be computationally costly. In the next, we describe our method on forgetting targeted token sequences without re-training the original base LM.

\subsubsection{Leave-One-Out Ensemble Mechanism}
\label{sec:method:looe}
Given any token sequence $(w_1,w_2,\ldots,w_T)\in\mathcal{D}_k^{\text{pri}}$, we assume its subsequence $(w_{i_1},\ldots,w_{i_{T'}})\in\mathcal{D}_k^{\text{forget}}$ is requested to be forgotten by the $k$th user, where $\mathcal{F}=\{i_1, \ldots, i_{T'}\}$ denotes their indices.

For any teacher LM $\theta_m$, it conducts the inference on this targeted sequence $(w_1,w_2,\ldots,w_T)$ and outputs probability distribution $p_{\theta_m}(
\cdot|w_{1:t-1})$ over all words in the vocabulary at step $t$. We utilize the ensemble mechanism where at each step $t$, the predicted probabilities from teachers are averaged on each word in the vocabulary. However, since we need to have this targeted sequence being forgotten by any updated model, the teacher model that was previously trained on the corpus containing the data from the $k$th user will be left out in such aggregation. Specifically, the following ensemble teacher output is used to supervise a student model at step $t$
\begin{align}
g_{-k}^{\text{LOO-E}}(\cdot|w_{1:t-1})=\frac{1}{M-1}\sum_{m=1}^M \textbf{1}\{\mathcal{D}_k \not\subset \mathcal{B}_m\}\cdot p_{\theta_m}(\cdot|w_{1:t-1})
\end{align}

Given that, consider the following loss function when training a student model on the targeted sequence $(w_1,w_2,\ldots,w_T)\in\mathcal{D}_k^{\text{pri}}$
\begin{align}
\label{loss}
\mathcal{L}(\theta)&=\sum_{t=1}^T\textbf{1}\{t \in \mathcal{F}\}\cdot \mathcal{L}_t^{\text{KL}}(\theta) \\
\mathcal{L}^{\text{KL}}_t(\theta):&=D_\text{KL}(g_{- k}^{\text{LOO-E}}(\cdot|w_{1:t-1})\,||\,p_{\theta}(\cdot|w_{1:t-1})) \label{loss:kl}
\end{align}
where $D_\text{KL}(P\,||\,Q)$ represents the Kullback–Leibler divergence between distributions $P$ and $Q$.

Given the corpus $\mathcal{D}^{\text{forget}}$ to be forgotten, we \emph{fine-tune} the original LM $\theta^{\text{base}}$ on this corpus using the loss above. Once it is trained, we will update the LM with this student model for delivering to users.

\subsection{Model Ensemble versus Data Ensemble}
In the framework described above, we ensemble the predicted probabilities from leave-one-out teacher models at each step. When the size of training corpus is small but $M$ is large, the distributional difference between $p_{\theta^*}(\cdot|w_{1:t-1})$ and $g_{-k}^{\text{LOO-E}}(\cdot|w_{1:t-1})$ might be no longer negligible on any sequence $(w_1,w_2,\ldots,w_T)\in\mathcal{D}_k^{\text{pri}}$. In that case, the ensemble can be performed at the data level instead where the $m'$th teacher trains on the corpus of $\cup_{m=1, m\neq m'}^{M} \mathcal{B}_m$. Then it can be directly used for supervising a student model on any sequences to be forgotten from $ \mathcal{D}_k^{\text{pri}}$.

Although data ensembling could lead to better estimation of $p_{\theta^*}$ under data scarcity, it requires $M-1$ times more computational cost compared with the model ensemble approach.

\subsection{The Option of Adding Gaussian Noise}
Building on top of the leave-one-out ensemble framework described above, random noises can be added to the aggregated outputs from teacher LMs so that they can further mask the presence of sensitive sequences. Here, we leverage the Gaussian mechanism which adds noise independently sampled from a Gaussian distribution $\mathcal{N}(0,\sigma^2)$ to every coordinate of the predicted probabilities, after which the re-normalization over vocabulary space is needed. Specifically, the teacher supervision $\mathcal{L}^{\text{KL}}_t(\theta)$ in~(\ref{loss:kl}) can be adjusted as
\begin{align}
D_\text{KL}(s(g_{- k}^{\text{LOO-E}}(\cdot|w_{1:t-1})+\mathcal{N}(0,\sigma^2))\,||\,p_{\theta}(\cdot|w_{1:t-1}))
\end{align}
where $s(\cdot)$ is a normalization function over the vocabulary such that all probabilities with added noises are truncated to non-negative and their sum equals to 1 after normalization.

\section{Theoretical Analysis}
\label{sec:the}
We study theoretical properties about the asymptotic equivalence of the predicted probabilities between leave-one-out teacher ensembles and the re-trained LM with all targeted private sequences excluded. Specifically, for any token sequence in $\mathcal{D}_k$, we show that the predicted probabilities between $g_{-k}^{\text{LOO-E}}$ and $\theta^{*}$ are negligible under certain conditions.

For simplicity, suppose each user contains the same number of training records where $|\mathcal{D}_k^{\text{pri}}|=n_{p}$ and $|\mathcal{D}_k^{\text{pub}}|=n$ for any $k$. For each record, we assume its token sequence has a fixed length of $T$. Additionally, there is no overlap of any private data across different users, that is, $\mathcal{D}_{k'}^{\text{pri}}\cap\mathcal{D}_{k''}^{\text{pri}}=\emptyset$ for any $k'\neq k''$. The public data of any user comes from the same distribution, and for the sake of simplicity, we assume $\mathcal{D}_{k'}^{\text{pub}}=\mathcal{D}_{k''}^{\text{pub}}$ for any $k'\neq k''$. We further assume $\mathcal{D}_k^{\text{forget}}=\mathcal{D}_k^{\text{pri}}$ for any $k$, thus users request to forget all their private information from the model. For any LM $\theta'$ trained on the corpus of $\mathcal{D}'$, we assume $p_{\theta'}(\textbf{w})=h(\sum_{\textbf{x} \in\mathcal{D}'}\textbf{1}\{\textbf{x}=\textbf{w}\}/|\mathcal{D}'|)$ in this analysis, where $h(\cdot)$ is a smoothing function. Intuitively, the predicted LM probability is based on the relative frequency count.
\begin{assumption}
\label{amp:lip}
The positive-valued function $h(\cdot)$ is Lipschitz continuous: there exists a positive real constant $C$ such that, for all real $x'$ and $x''$, it satisfies $|h(x')-h(x'')|\leq C\cdot |x'-x''|$.
\end{assumption}

\begin{theorem}
\label{thm:prob}
Suppose Assumption \ref{amp:lip} holds. Let $k^*$ be the index of user who requests that their information be forgotten. Suppose $n_p=o(n)$. Then for any token sequence $\textbf{\emph{w}}\in\mathcal{D}_{k^*}$, the leave-one-out teacher ensembles model satisfies $g_{-k^*}^{\emph{LOO-E}}(\textbf{\emph{w}})/p_{\theta^*}(\textbf{\emph{w}})\rightarrow 1$ as $n\rightarrow\infty$.
\end{theorem}
The proof of the theorem is deferred to Appendix.

\section{Experiments}
\label{sec:expt}

\subsection{Datasets}
Our experiments use the LibriSpeech extended text-only corpus \cite{openSLR11} for LM training purpose. It is collected from 14,500 public domain books which contains around 40M sentences.

For the evaluation of LM and ASR tasks, we utilize the test split of LibriSpeech ASR corpus and text transcripts \cite{panayotov2015librispeech}, which includes around 5.5K utterances from approximately 70 speakers.

We also experiment with the WikiText-103 corpus \cite{merity2016pointer}, where the train and test sets contain 1.5M and 15K sentences, respectively.

\subsection{Canaries}
To simulate rare or unique token sequences in LM training data, we build on the “secret sharer" framework in \cite{carlini2019secret}. Specifically, random textual sequences, called \emph{canaries}, are generated and then inserted into a training corpus. Here, canaries aim to mimic private user data.

The procedure of inserting canaries into any text set is described as follows: (1) randomly shuffle all the records, and create synthetic user IDs where each user owns 100 records; (2) randomly pick 100 users and for each user, a random 5-word canary is generated which simulates the ``private sequence'' from that specific user. Notice that each word in any generated canary is contained in LM vocabulary and no canary is shared by different users; (3) each generated canary is inserted into a training corpus for 100 times being repeated.

When users request to delete these canaries from a trained LM on the corpus with inserted canaries, we compare multiple methods and measure the frequencies of having these canaries forgotten.

We use the following two techniques to evaluate the privacy risks on extracting canaries from an LM:
\begin{itemize}
  \item Beam Search (BS). We utilize beam search (beam size is 100) to check if any canary is contained in the top 100 most-likely 3-word continuations from the 2-word prefix of the canary;
  \item Random Sampling (RS). We say any canary is extracted from an LM if it has the least perplexity (PPL) among 1,000 random suffixes, given the 2-word prefix of the canary.
\end{itemize}
In our experiments, we report the frequencies of times that the 100 generated canaries are detected by BS or RS in any LM.

\subsection{Setups}
The LM evaluated in our experiments is LSTM based with embeddings dimension 300, and 2 layers of 1,500 hidden units. The word vocabulary set is around 10K.

In the experiments, we consider the following approaches:
\begin{itemize}
  \item \texttt{Baseline} LM trains on the corpus with canaries, which is the original base model deployed to users;
  \item \texttt{Re-Trained} LM trains on the corpus with all canaries being deleted, when there are requests to remove data;
  \item \texttt{PATE} represents the privacy-preserving teacher aggregation based LM training method in \cite{papernot2016semi};
  \item \texttt{GA} refers to the gradient ascent method in \cite{jang2022knowledge} which applies post-training to forget targeted canaries upon the receipt of requests to remove data;
  \item \texttt{LOO-E} is the proposed approach for post-training unlearning of canaries when there are requests to remove data.
\end{itemize}
In each of the \texttt{Baseline}, \texttt{Re-Trained}, and \texttt{PATE} methods, the LMs are trained for 10 epochs (denoted as \texttt{EP}); in \texttt{PATE} and \texttt{LOO-E} methods, 5 teachers (denoted as \texttt{5T}) are used and also trained for 10 epochs; during the post-training unlearning (fine-tuning), learning rates (denoted as $r$) and numbers of epochs are varied for \texttt{GA} and \texttt{LOO-E} methods in the experiments.

On the test split of LibriSpeech ASR corpus, we also evaluate the ASR performance, in terms of word-error-rate (WER), with the LMs being used as second-pass rescorers on the generated 20-best hypotheses. The ASR model is a RNN-T model with the Emformer encoder \cite{emformer2021streaming}, LSTM predictor, and a joiner. It has around 80 million parameters and is trained from scratch using the train split of LibriSpeech ASR corpus (without canaries).

\subsection{Results}
Table~\ref{tab:main} shows the perplexity results on LibriSpeech test split as well as the percentages of canaries being uncovered by BS and RS techniques. From the results, we have the following observations:
\begin{itemize}
    \item Compared with the \texttt{Baseline}, the \texttt{Re-Trained} method re-trains on the corpus with all canaries removed and is thus able to prevent them from being detected by BS or RS while keeping the perplexity impact small. However, it suffers from high computational costs since the model has to be re-trained every time users request that their information be deleted;
    \item \texttt{PATE} is considered as a generic training approach to reduce the memorization for any rare token sequences, and it does have some effect on hiding these canaries. However, there are still many canaries that can be extracted by BS or RS while the perplexity is also slightly lifted;
    \item \texttt{GA} performs gradient ascent based LM fine-tuning on these targeted canaries. Some of the training configurations leads to full forgetting of canaries but perplexity is greatly degraded, while others have less increased perplexity but still have quite a few canaries being detected. In general, larger learning rates or more fine-tuning epochs result in worse perplexity but better mitigation performance on having canaries extracted;
    \item The proposed \texttt{LOO-E} achieves the best privacy-utility trade-offs compared with \texttt{PATE} and \texttt{GA}. Specifically, the increase of perplexity is negligible, while almost all the canaries can be forgotten by this approach as in the model of L3. Similar to \texttt{GA} as a post-training unlearning method, more fine-tuning epochs leads to higher perplexity but less canaries detected.
\end{itemize}

\begin{table}[ht!]
  \vspace{-0.4cm}
  \caption{Results on LibriSpeech ASR test corpus, where PPLs and percentages of canaries detected by BS and RS are reported.}
  \centering
  \resizebox{0.9\columnwidth}{!}{%
  \begin{threeparttable}
  \begin{tabular}{c|l|r|rr}
    \toprule
    \multicolumn{2}{c|}{\emph{}} & \emph{Utility}
    & \multicolumn{2}{|c}{\emph{Privacy}} \\
    \cmidrule{1-2} \cmidrule{3-5}
    \emph{ID} & \emph{Method} & \emph{PPL} & \emph{BS\,\,} & \emph{RS\,\,} \\
    \midrule
    B1 & \texttt{Baseline} & 71.7 & 100\% & 100\% \\
    \midrule 
    R1 & \texttt{Re-Trained} & 72.5 & 0\% & 0\% \\
    \midrule 
    P1 & \texttt{PATE(5T)} & 74.9 & 2\% & 74\% \\
    \midrule
    G1 & \texttt{GA($r=4e^{-5}$,1EP)} & 83.4 & 0\% & 8\% \\
    G2 & \texttt{GA($r=1e^{-5}$,1EP)} & 72.8 & 51\% & 100\% \\
    G3 & \texttt{GA($r=1e^{-5}$,3EP)} & 80.5 & 0\% & 16\% \\
    G4 & \texttt{GA($r=4e^{-6}$,5EP)} & 76.8 & 0\% & 63\% \\
    G5 & \texttt{GA($r=4e^{-6}$,10EP)} & 94.2 & 0\% & 1\% \\
    G6 & \texttt{GA($r=1e^{-6}$,10EP)} & 73.0 & 8\% & 100\% \\
    G7 & \texttt{GA($r=1e^{-6}$,20EP)} & 77.0 & 0\% & 56\% \\
    \midrule
    L1 & \texttt{LOO-E(5T,$r=4e^{-5}$,5EP)} & 72.5 & 0\% & 30\% \\
    L2 & \texttt{LOO-E(5T,$r=4e^{-5}$,10EP)} & 72.7 & 0\% & 13\% \\
    L3 & \texttt{LOO-E(5T,$r=4e^{-5}$,20EP)} & 73.1 & 0\% & 5\% \\
    \bottomrule
  \end{tabular}
  \end{threeparttable}
  }
  \vspace{-0.2cm}
  \label{tab:main}
\end{table}

The computational cost for fine-tuning on these canaries is less than 0.1\% of that for training on the entire LibriSpeech text-only corpus. Thus, suppose users submit data removal requests for $v$ times, then the computational costs of the \texttt{Re-Trained} method and the proposed \texttt{LOO-E} method are estimated as $U\cdot v$ and $U \cdot (1 + 0.001v)$, respectively. Here, the latter takes into account the costs of training teacher models and $U$ is the unit cost for training on the full corpus.

Furthermore, Tables~\ref{tab:extra}-\ref{tab:noise} evaluate the effect of leave-one-out data ensemble method (denoted as \texttt{LOO-D-E}), freezing the LM embedding layers during the fine-tuning (denoted as \texttt{F-Embed}), as well as adding Gaussian noises to the teachers' outputs with hyperparameter $\sigma$ controlling the magnitude of noises. From the results, we find that the performance of models D1-D3 and L4-L6 is comparable with that of L1-L3 in term of the privacy-utility trade-offs. By adding small Gaussian noises with $\sigma=1e^{-5}$ to the ensembles, model L9 slightly outperforms model L3. Also, it is interesting to notice that adding small Gaussian noises also improves the \texttt{PATE} method in the comparison of model P2 with P1.

\begin{table}[ht!]
  \vspace{-0.4cm}
  \caption{LibriSpeech results on data ensemble and layer freezing.}
  \centering
  \resizebox{\columnwidth}{!}{%
  \begin{threeparttable}
  \begin{tabular}{c|l|r|rr}
    \toprule
    \multicolumn{2}{c|}{\emph{}} & \emph{Utility}
    & \multicolumn{2}{|c}{\emph{Privacy}} \\
    \cmidrule{1-2} \cmidrule{3-5}
    \emph{ID} & \emph{Method} & \emph{PPL} & \emph{BS\,\,} & \emph{RS\,\,} \\
    \midrule
    D1 & \texttt{LOO-D-E(5T,$r=4e^{-5}$,5EP)} & 72.8 & 0\% & 42\% \\
    D2 & \texttt{LOO-D-E(5T,$r=4e^{-5}$,10EP)} & 73.0 & 0\% & 15\% \\
    D3 & \texttt{LOO-D-E(5T,$r=4e^{-5}$,20EP)} & 73.3 & 0\% & 3\% \\
    \midrule
    L4 & \texttt{LOO-E(5T,$r=4e^{-5}$,5EP,F-Embed)} & 72.5 & 0\% & 35\% \\
    L5 & \texttt{LOO-E(5T,$r=4e^{-5}$,10EP,F-Embed)} & 72.7 & 0\% & 14\% \\
    L6 & \texttt{LOO-E(5T,$r=4e^{-5}$,20EP,F-Embed)} & 73.1 & 0\% & 3\% \\ 
    \bottomrule
  \end{tabular}
  \end{threeparttable}
  }
  \vspace{-0.2cm}
  \label{tab:extra}
\end{table}

\begin{table}[ht!]
  \caption{LibriSpeech results on the effect of adding Gaussian noises.}
  \centering
  \resizebox{\columnwidth}{!}{%
  \begin{threeparttable}
  \begin{tabular}{c|l|r|rr}
    \toprule
    \multicolumn{2}{c|}{\emph{}} & \emph{Utility}
    & \multicolumn{2}{|c}{\emph{Privacy}} \\
    \cmidrule{1-2} \cmidrule{3-5}
    \emph{ID} & \emph{Method} & \emph{PPL} & \emph{BS\,\,} & \emph{RS\,\,} \\
    \midrule
    P2 & \texttt{PATE(5T,$\sigma=1e^{-5})$} & 74.6 & 0\% & 24\% \\
    P3 & \texttt{PATE(5T,$\sigma=1e^{-4})$} & 90.6 & 0\% & 7\% \\
    P4 & \texttt{PATE(5T,$\sigma=1e^{-3})$} & 261.2 & 0\% & 2\% \\
    \midrule 
    L7 & \texttt{LOO-E(5T,$r=4e^{-5}$,5EP,$\sigma=1e^{-5}$)} & 71.7 & 0\% & 34\% \\
    L8 & \texttt{LOO-E(5T,$r=4e^{-5}$,10EP,$\sigma=1e^{-5}$)} & 71.7 & 0\% & 14\%  \\
    L9 & \texttt{LOO-E(5T,$r=4e^{-5}$,20EP,$\sigma=1e^{-5}$)} & 72.3 & 0\% & 0\% \\
    \bottomrule
  \end{tabular}
  \end{threeparttable}
  }
  \label{tab:noise}
\end{table}

\begin{table}[ht]
  \vspace{-0.4cm}
  \caption{WER results on LibriSpeech ASR test corpus as well as the synthetic audio set generated using TTS on the canaries.}
  \centering
  \resizebox{\columnwidth}{!}{%
  \begin{threeparttable}
  \begin{tabular}{l|c|c|c|c}
    \toprule
    \multicolumn{2}{c|}{\emph{}}
    & \multicolumn{3}{|c}{\emph{ASR w/ 2nd-Pass LM Rescore}} \\
    \cmidrule(r){1-5}    
    \emph{Evaluation Dataset} & \emph{ASR (w/o LM)} & $\quad$B1$\quad$ & $\quad$R1$\quad$ & L9 \\
    \midrule
    LibriSpeech Test Split & 6.17 & 5.86 & 5.87 & 5.89 \\
    \midrule
    TTS-ed Canaries$^*$ & 30.06 & 23.14 & 30.06 & 30.06 \\
    \bottomrule
    \multicolumn{5}{l}{$^*$Lower WER indicates more exposure of private canaries in the ASR model.}
  \end{tabular}
  \end{threeparttable}
  }
  \vspace{-0.4cm}
  \label{tab:wer}
\end{table}

Table~\ref{tab:wer} shows the WER results on LibriSpeech ASR test corpus with second-pass LM rescoring. Compared with the \texttt{Baseline} and \texttt{Re-Trained} methods, WER of the proposed approach L9 is not compromised. It is also interesting to study to what extent the ASR models are exposed to privacy risks when LMs are used as rescorers. We use the text-to-speech (TTS) technique to generate a synthetic audio set according to these canaries and test how accurately the ASR model with different LMs can recognize the tokens in canaries. Here, better recognition indicates more privacy risks. Seen from the results in Table~\ref{tab:wer}, WER from model L9 is not lower than the one without LM or the \texttt{Re-Trained} LM, which means all canaries are properly forgotten by this LM.

Besides the results on LibriSpeech above, we experiment with WikiText-103 where the training corpus is much smaller in size compared with the LibriSpeech text-only corpus. Table~\ref{tab:wiki} shows the comparison results over these methods. The observations are similar to what we find on LibriSpeech experiments.

\begin{table}[ht!]
  \vspace{-0.4cm}
  \caption{Results on WikiText-103 test corpus.}
  \centering
  \resizebox{0.9\columnwidth}{!}{%
  \begin{threeparttable}
  \begin{tabular}{c|l|r|rr}
    \toprule
    \multicolumn{2}{c|}{\emph{}} & \emph{Utility}
    & \multicolumn{2}{|c}{\emph{Privacy}} \\
    \cmidrule{1-2} \cmidrule{3-5}
    \emph{ID} & \emph{Method} & \emph{PPL} & \emph{BS\,\,} & \emph{RS\,\,} \\
    \midrule
    B1 & \texttt{Baseline} & 66.1 & 100\% & 100\% \\
    \midrule 
    R1 & \texttt{Re-Trained} & 66.0 & 0\% & 0\% \\
    \midrule 
    P1 & \texttt{PATE(5T)} & 70.3 & 39\% & 100\% \\
    \midrule
    G2 & \texttt{GA($r=1e^{-5}$,1EP)} & 71.1 & 86\% & 100\% \\
    G3 & \texttt{GA($r=1e^{-5}$,3EP)} & 95.9 & 0\% & 56\% \\
    G5 & \texttt{GA($r=4e^{-6}$,5EP)} & 117.2 & 0\% & 21\% \\
    \midrule
    L1 & \texttt{LOO-E(5T,$r=4e^{-5}$,5EP)} & 69.0 & 0\% & 13\% \\
    D1 & \texttt{LOO-D-E(5T,$r=4e^{-5}$,5EP)} & 68.5 & 0\% & 15\% \\
    \bottomrule
  \end{tabular}
  \end{threeparttable}
  }
  \vspace{-0.2cm}
  \label{tab:wiki}
\end{table}

\section{Conclusion}
\label{sec:con}
In this work, we propose a new leave-one-out ensemble approach to unlearn the targeted textual sequences that are requested to be forgotten by the model. Experiments on LibriSpeech and WikiText-103 datasets show that the presented approach achieves superior privacy-utility trade-offs than other counterparts.

Future work might include experiments with large LMs \cite{touvron2023llama2} as well as the parameter-efficient fine-tuning techniques \cite{hu2021lora} during the post-training unlearning process for the proposed method.

\bibliographystyle{IEEEbib}
\bibliography{refs}


\clearpage
\section{Appendix}
\label{appen}
\subsection{Proof of Theorem \ref{thm:prob}}
\emph{Proof}. For any token sequence $\textbf{w}\in\mathcal{D}_{k^*}^{\text{pri}}$, it is straightforward to see that $p_{\theta^*}(\textbf{w})=g_{-k^*}^{\text{LOO-E}}(\textbf{w})=h(0)$ since this sequence was excluded in the training records for both. 

For any $\textbf{w}\in\mathcal{D}_{k^*}^{\text{pub}}$, denote
\begin{align}
l:=\sum_{\textbf{x}\in\mathcal{D}_{k^*}^{\text{pub}}}\mathbf{1}\{\textbf{x}=\textbf{w}\}
\end{align}
and then we have $p_{\theta^*}(\textbf{w})=h(l/n)$ given $\mathcal{D}_{k'}^{\text{pub}}=\mathcal{D}_{k''}^{\text{pub}}$ for any $k'\neq k''$. Also, $g_{-k^*}^{\text{LOO-E}}(\textbf{w})=h(l/(n + n_p))$. Then
\begin{align}
&|\log(p_{\theta^*}(\textbf{w}))-\log(g_{-k^*}^{\text{LOO-E}}(\textbf{w}))| \\
&=|\log(h(l/n))-\log(h(l/(n + n_p)))| \\
\label{eq:lip}
&\leq C \cdot |l/n - l/(n + n_p)| \\
&= C \cdot l/n \cdot \bigg|1-\frac{1}{1+n_p/n}\bigg | \\
\label{eq:conv}
&\rightarrow 0
\end{align}
Here, (\ref{eq:lip}) follows from Assumption \ref{amp:lip}; (\ref{eq:conv}) is implied from $l=O(n)$ and $n_p=o(n)$.


\end{document}